\documentclass[pmlr]{jmlr}% new name PMLR (Proceedings of Machine Learning)

\usepackage{longtable}% for long tables

\usepackage[singlelinecheck=off]{caption}

 \usepackage{booktabs}
 
\usepackage[load-configurations=version-1]{siunitx} % newer version

\makeatletter
\def\set@curr@file#1{\def\@curr@file{#1}} %temp workaround for 2019 latex release
\makeatother

\theorembodyfont{\upshape}
\theoremheaderfont{\scshape}
\theorempostheader{:}
\theoremsep{\newline}

\title[CXR-RePaiR-Gen]{Retrieval Augmented Chest X-Ray Report Generation using OpenAI GPT models
}

\author{\Name{Mercy Ranjit}\thanks{MR is also associated with Microsoft Research, Bangalore}
       \Email{meranjit@microsoft.com}\\ 
       \addr Department of Computer Science\\
       Bharathidasan University, \\
       Trichy, India 
       \AND
       \Name{Gopinath Ganapathy}
       \Email{gganapathy@gmail.com}\\ 
       \addr Department of Computer Science\\
       Bharathidasan University,\\
       Trichy, India
       \AND
       \Name{Ranjit Manuel}\thanks{RM  is also associated with Databricks Inc., Bangalore}
       \Email{ranjit.f.manuel@gmail.com}\\ 
       \addr AI for Digital Health and Imaging\\
       Indian Institute of Science,\\
       Bengaluru, India
       \AND
       \Name{Tanuja Ganu}
       \Email{taganu@microsoft.com}\\ 
       \addr Microsoft Research,\\
       Bengaluru, India} 

\editor{Editor's name}

\begin{document}

\maketitle

\begin{abstract}
We propose Retrieval Augmented Generation (RAG) as an approach for automated radiology report writing that leverages multimodally aligned embeddings from a contrastively pretrained vision language model for retrieval of relevant candidate radiology text for an input radiology image and a general domain generative model like OpenAI \texttt{text-davinci-003}, \texttt{gpt-3.5-turbo} and \texttt{gpt-4} for report generation using the relevant radiology text retrieved. This approach keeps hallucinated generations under check and provides capabilities to generate report content in the format we desire leveraging the instruction following capabilities of these generative models. Our approach achieves better clinical metrics with a BERTScore of 0.2865 ($\Delta$+ 25.88 \%) and $S_{emb}$ score of 0.4026 ($\Delta$+ 6.31 \%). Our approach can be broadly relevant for different clinical settings as it allows to augment the automated radiology report generation process with content relevant for that setting while also having the ability to inject user intents and requirements in the prompts as part of the report generation process to modulate the content and format of the generated reports as applicable for that clinical setting. 
\end{abstract}

\section{Introduction}
Automated Radiology Report Generation Systems can improve the report writing workflow of radiologists in various ways. These AI systems can generate free text content or structured report content for review by the radiologists w.r.t various attributes of interests like organ systems, pathology, abnormalities, severity, size or location of findings, progression status etc. 

Some of the existing work around AI enabled radiology report generation cast the radiology report generation problem as an image captioning problem or generative task [\cite{chen2020generating,miura2020improving}].   An interesting recent work CXR-RePaiR [\cite{endo2021retrieval}] cast it as a retrieval problem taking advantage of the finite set of diagnostic details and attributes associated with radiology images. This approach is powerful in that it can leverage a very large database of past and present radiological reports while making impression recommendations. But all these approaches suffer from various issues related to irrelevant content or hallucinations in the generations.  

A very recent work CXR-ReDonE by \cite{ramesh2022improving} addressed one such issue in report generation related to prior references by creating a new dataset CXR-PRO [\cite{rameshcxr}] that eliminated the prior references found in the radiology text reports of MIMIC-CXR [\cite{johnson2019mimic}]  and used this dataset to improve CXR-RePaiR [\cite{endo2021retrieval}] to establish a new SOTA benchmark in radiology report impression generations.

But purely retrieval-based report writing models like CXR-RepaiR and CXR-ReDonE does not allow the user to modulate the retrievals per their need which is useful for applicability in various clinical settings. Below are some limitations with pure retrieval-based approaches, some of which were mentioned for future work in these papers:

\begin{itemize}
\item	Pure retrieval-based report generation may include irrelevant retrievals for no-findings case reports especially when the number of relevant retrievals, K is set $>$ 1 for generating the report. 
\item	There are some unwanted noises in the generations like prior report references, mention of doctor names, user specific details, etc which are extracted as-is if they are available. Pure retrieval-based report generation also suffer from duplicate content in the retrievals.
\item	It is also not possible to generate the radiology report in a specific format which may be the requirement for different report generation downstream applications. It may be useful to create structured radiology reports extracting attributes of interest like pathologies, positional information, severity, size etc instead of a raw text output.
\item	Pure retrieval-based systems may have some incoherent information in the generated report as the retrievals may bring sentences from reports belonging to two different patients as-is.
\end{itemize}

With very capable generative Large Language Models (LLM) like \texttt{text-davinci-003},  \texttt{gpt-3.5-turbo}(Chat GPT) and  \texttt{gpt-4} being available for the general domain which can generate relevant content based on instructive prompts in a zero-shot or few-shot setting for a wide variety of downstream tasks, it would be useful to explore how they can be leveraged for the work of radiology report generation to assist the radiologists. These models, however, lack the up-to-date information or domain specific information required specifically in a medical domain setting.  Updated and relevant domain specific content when available for these models during the time of generation can allow to extend the capabilities of these large language models to do tasks with data, they were not exposed to during the training phase. In-addition we can leverage the instructions following capabilities of these models to elicit the required responses we require from these models. This additional context available to the LLMs for generations makes them hallucinate less. We are motivated by the advantages of Retrieval Augmented Generation (RAG) experimented in the work by \cite{lewis2020retrieval} which showed that the generations from RAG are more strongly grounded in real factual knowledge causing less hallucinations and its broad application for various downstream tasks called out in the “Broader Impact” section of the paper. 

\begin{figure}
    \centering
    \includegraphics[width=1.0\linewidth, ]{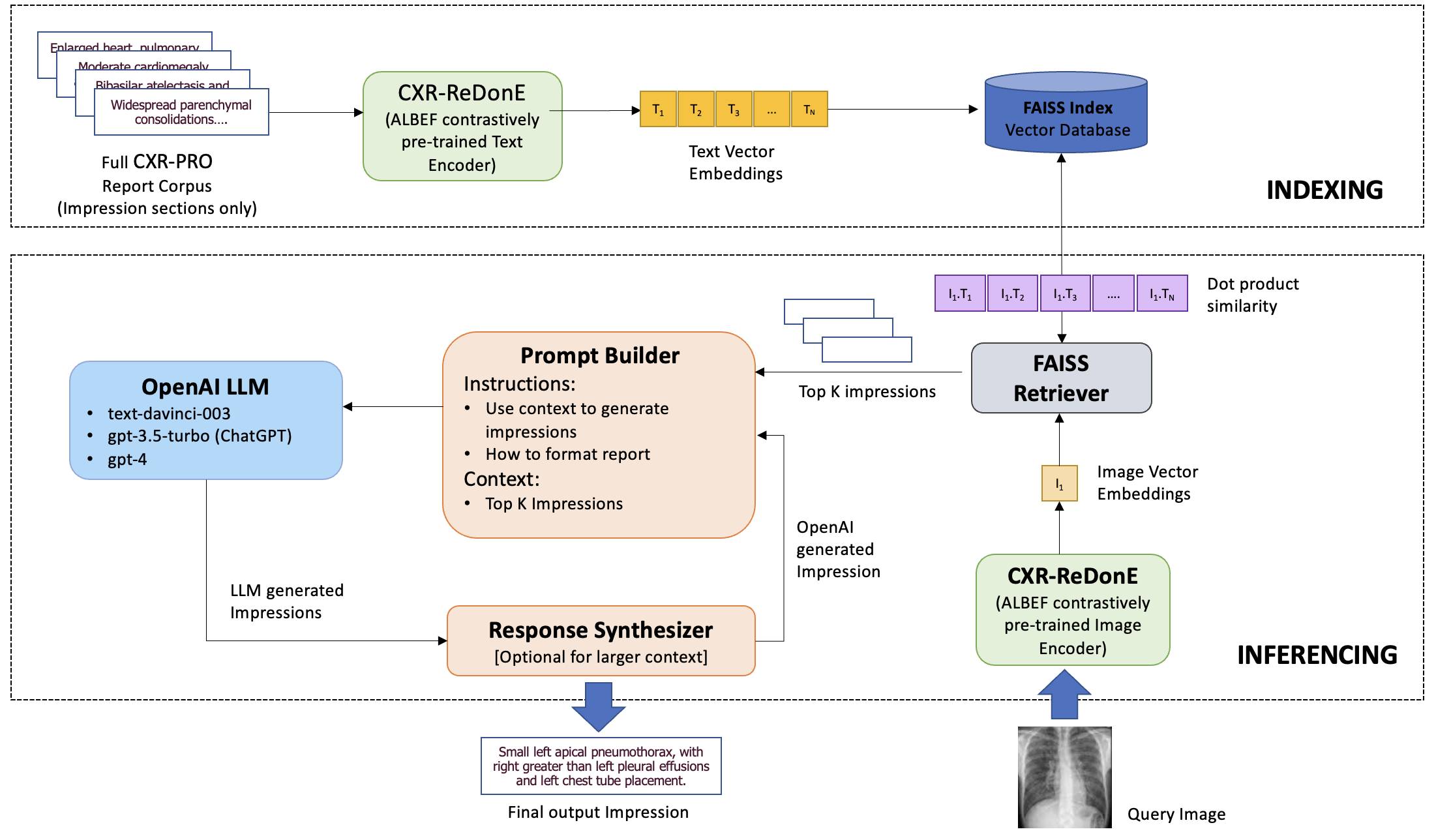}
    \caption{We project all the text embeddings of sentences from radiology impression using a contrastively pretrained vision-language encoder (CXR-ReDonE) to a vector database index and retrieve the most matching sentences for an input image embedding using the same encoder model. The retrieved impression reports or sentences form the context of the prompt to the LLM along with instructions to generate the impression.}
    \label{fig:Figure 1}
\end{figure}

We propose Contrastive X-ray-Report Pair Retrieval based Generation (CXR-RePaiR-Gen) as a RAG based methodology for radiology report generation extending on top of the work by CXR-RepaiR [\cite{endo2021retrieval}] and CXR-ReDonE [\cite{ramesh2022improving}]. As illustrated in \autoref{fig:Figure 1}, we leverage the contrastively pretrained ALBEF model [\cite{li2021align}] from CXR-ReDonE to generate the vision-language aligned embeddings for a database of radiology reports. The same model is used to generate the embedding for an input radiology image. As the image and text embeddings were aligned during the contrastive pre-training, the most relevant text radiology text (reports or sentences) is retrieved for an input x-ray image based on the similarity of the input image embeddings to the radiology report embeddings. A consolidated radiology report impression is generated from the filtered set of records using the OpenAI text-davinci-003, gpt-3.5-turbo and gpt-4 models. 

RAG based approach not only makes the radiology report generations grounded on the relevant radiology text retrieved from the radiology text corpus but also allows the user to inject user intents as instructions and few shot examples as part of the generation process via prompt engineering to generate content in the required format applicable for the clinical setting. 

\subsection*{Generalizable Insights about Machine Learning in the Context of Healthcare}
Our approach brings the below key insights for ML in healthcare: 

\begin{itemize}
\item	Retrieval augmented generation can help bridge the advantages of various domain specific healthcare encoder models with that of the general domain generative models leveraging the best of both models. We show this improves the clinical metrics. 
\item	We also measure the radiology report generations for hallucinations by comparing the LLM generated response with the retrieved radiology text from the radiology reports or sentences corpus. This can help in decision making when planning to practically use these systems in a real clinical setting.
\item	Our paper also shows how we can leverage prompt engineering in LLM to inject user intents and requirements to produce radiology reports in different output formats relevant for the downstream application with few-shot learning. 
\end{itemize}

Our approach achieves better clinical metrics with a BERTScore [\cite{yu2022evaluating}] of 0.2865 ($\Delta$+ 25.88\%) and $S_{emb}$ score [\cite{endo2021retrieval}] of 0.4026 ($\Delta$+ 6.31\%) over the previous state-of-the-art retrieval method CXR-ReDonE. In clinical settings, the improvement of these scores means we are able to generate radiology reports that are closer to the ground truth impression semantically, at the same time being very concise reducing the noise from the retrievals. We are also on par with CXR-ReDonE on the RadGraph $F_{1}$ metric [\cite{yu2022evaluating}]. This metric measures if we are able to retrieve all the clinical entities accurately. We cannot exceed on this metric beyond the retrieval model CXR-ReDonE as the RAG generation are based on the retrieved records from CXR-ReDonE.

\section{Related Work}
Recent works in radiology report generation approached the problem as a generative task like the work of \cite{chen2020generating} which used a Transformer decoder architecture R2Gen and the work of \cite{miura2020improving} which focused on generating complete, consistent, and clinically accurate reports using a reward-based reinforcement leaning approach by name M2 Trans.

\cite{endo2021retrieval} in their work CXR-RePaiR approached radiology report generation problem using a retrieval approach and set a new SOTA benchmark using more clinically reliable metrics.  The retrieval was based on their constrastively pretrained vision-language model using the MIMIC-CXR dataset [\cite{johnson2019mimic}].  A new clinical efficacy similarity metric $S_{emb}$ was introduced in the paper to calculate the semantic similarity using the last hidden representations from the CheXbert [\cite{smit2020chexbert}] labeler between the reference report and the predicted report. The paper also used the BERTScore metric [\cite{zhang2019bertscore}] as another measure for semantic similarity. 

\cite{ramesh2022improving} in their work addressed one key issue pertaining to all automated radiology report generation approaches which are prior report references in the radiology report which impacts the quality of report generation. They built a new dataset CXR-PRO [\cite{rameshcxr}] by addressing this issue on the MIMIC-CXR dataset [\cite{johnson2019mimic}]. They also retrained CXR-RepaiR using the CXR-PRO dataset and an updated architecture ALBEF [\cite{li2021align}] and set the current SOTA for the radiology report generation task. They also used the RadGraph F1 [\cite{yu2022evaluating}] score as an additional metric to measure the completeness and accuracy of the clinical entities available in the predicted report using the RadGraph model [\cite{jain2021radgraph}]

With the rise of the LLMs, Retrieval Augmented Generation (RAG) was introduced in the work by \cite{lewis2020retrieval} which brought some key advantages of leveraging external knowledge sources to augment the knowledge of LLMs to do a task. LLMs generations are also strongly grounded in real factual knowledge which makes it “hallucinate” less and produce generations that are more factual. The broarder impact statement from the paper mentioned its application in a wide variety of scenarios, for example by endowing it with a medical index.

We in this work endow the LLMs with the index of radiology report text and use it as a knowledge base to allow LLMs generate a radiology report impression for an input radiology image. To enable the multimodal retrieval of Images and Text, we use the constrastively pretrained vision language model from CXR-ReDonE [\cite{ramesh2022improving}] which improved the work of CXR-RepaiR [\cite{endo2021retrieval}] for multimodal retrievals to see if augmented generation on top of these retrievals can further push the report generation benchmark. We also see if we can use the general capabilities of LLMs to modulate the report generation outputs per user requirements to enable a wide variety of usage patterns across different clinical settings. We also measure the hallucinations from RAG to pivot the application of the proposed approach in a real-world clinical setting.

\section{Methods}
In this paper we propose radiology report generation as a data augmented generation task using large language models like \texttt{text-davinci-003}, \texttt{gpt-3.5-turbo} and \texttt{gpt-4}.  Our hypothesis is that it is not required to have domain specific generative models, but domain specific retrievers. We can leverage embeddings from domain specific encoders for the data retrieval task and use the retrieved data for augmentation using a general domain generative model. We build on top of the work by CXR-RePaiR [\cite{endo2021retrieval}] and CXR-ReDonE [\cite{ramesh2022improving}] by considering the problem of radiology report generation as retrieval task but adding a generative step using the retrieved impressions. We use the contrastively pretrained vision language model ALBEF [\cite{li2021align}] from CXR-ReDonE to generate the image and text embeddings from the radiology images and reports. For selecting the top K records for data augmentations, we tried both report level corpus $R = {r_1, ..., r_n}$ and sentence level corpus $S = {s_1, ..., s_n}$ as in CXR-RePaiR. 

We use the ALBEF [\cite{li2021align}] model from CXR-ReDonE to generate contrastively aligned text embeddings for the radiology report text corpus using the CXR-PRO dataset [\cite{rameshcxr}] and index it in the vector database. We use the same model for generating the contrastively aligned image embedding for the input radiology image x and use the embedding to retrieve the top-K records from the reports or sentences corpus based on dot-product similarity.  The top-K reports or sentences that have the highest similarity to the input image embeddings are selected for augmenting the generation using \texttt{text-davinci-003}, \texttt{gpt-3.5-turbo} and \texttt{gpt-4} models. We generate impression I by prompting the LLM with prompt P passing in the retrieved top-K sentences of the sentence corpus S as the context along with instructions Q for the generation. 

$$
I=LLM(P(Q,\sum_{i=1}^kS_{i} ))
$$

Where top K sentences from S is selected using the function $ argmax_{ s \in S(R)}, f (s, x)$, f indicating the similarity dot product function on the input sentence s and radiology image x as in CXR-RePaiR.

If the size of the context records for the LLM generation is beyond the token limit of the LLM, we propose to use the refine methodology for iteratively generating the response from the LLM. We generate an initial impression from the LLM by prompting the LLM with the first retrieved report or sentence of the top K records from the radiology text corpus as the context and then we pass the initial impression response generated by the LLM, second sentence or report as context and so on to get a final refined answer from the LLM. $I_k$ denotes the iteratively constructed impression from the LLM using refine prompt $P^r$ constructed using the impressions from the previous iterations, instructions Q and $S_{k}$ denoting the kth sentence from the sentence corpus.

$$
I_k=LLM(P^r (Q,I_{k-1},S_k))
$$

Refine mechanism is available with LLM based frameworks like langchain [\cite{langchain2022}] and llama-index [\cite{llamaindex2022}]. As the size of context for RAG based generations could vary in different clinical settings, the paper proposes the usage of refine approach. In the context of our experiments, for this paper refine mechanism was not required as we experimented with smaller $K = {1,2,3}$ values and the context were still within the limit of LLMs and iterative response building was not required.

In addition to free text radiology report generation, we also hypothesize that it would be more useful to have the radiology reports in a structured format including key attributes of interest from the retrievals. These attributes of interest could be pathologies, severity related to pathology, severity, size, or position etc. We use prompt engineering with few shot examples to generate a structured radiology report output.

\subsection{Retrieval Corpus}

We base the retrieval corpus on the train impressions of the CXR-PRO dataset [\cite{rameshcxr}] which consists of 374,139 free-text radiology reports and their associated chest radiographs. As CXR-PRO is based on MIMIC-CXR which is a de-identified dataset, no protected health information (PHI) is included. CXR-PRO is an adapted version of the MIMIC-CXR dataset [\cite{johnson2019mimic}]] with prior references omitted.  It addresses the issue of hallucinated reference to priors produced by radiology report generation models. We use the impressions sections of the radiology reports in the corpus and consider both report-level impressions and as well as the sentences comprising the report-level impressions as the retrieval corpus for report generation as in CXR-RePaiR [\cite{endo2021retrieval}]

\subsection{Baselines}

We consider CXR-ReDonE [\cite{ramesh2022improving}] which does retrieval-based report generation with 
CXR-PRO dataset [\cite{rameshcxr}] as the retrieval corpus as our baseline. We aim to see if retrieval augmented generation using LLMs on top of these retrievals can help improve radiology report generation clinical metrics.

\subsection{Prompt Design}

We design two sets of prompts to generate the radiology report as free-text report, one for the \texttt{text-davinci-003} model and another for report generation in the conversational setup with the \texttt{gpt-3.5-turbo} and \texttt{gpt-4} model as shown in \autoref{tab:Table 1} We provide instructions to the LLM to use the retrieved sentences as a context to generate the radiology report. For \texttt{gpt-3.5-turbo} and \texttt{gpt-4} the prompt design involves system prompt and a user prompt in a conversation setting. The system prompt instructs the system to generate a radiology report impression from the context that the user will send. The user prompt sends the retrieved records as a context requesting the system to provide the radiology report as a response. 

\begin{table}[]
  \centering 
  \caption{Prompts for the OpenAI LLMs for Radiology Report Impression Generation from the retrieved reports as context text in the zero-shot setting. The text in italics in the prompts corresponds to the variables used in formatting the prompt.}
  \begin{tabular}{p{2in} p{2in} p{2in}}
  \toprule
    \textbf{text-davinci-003} & \textbf{gpt-3.5-turbo/gpt-4 \newline System Prompt} & \textbf{gpt-3.5-turbo/gpt-4 \newline User Prompt} \\
    \midrule
    Generate an impression summary for the radiology report using the context given. \newline 
    \newline
    Strictly follow the instructions below while generating the impressions. 
    \newline
    \newline
    \textcolor{red}{Instructions:} 
    \begin{itemize}
        \item Impression summary should be based on the information in the context.
        \item Limit the generation to \textit{maxlen} words.
    \end{itemize}
    \textcolor{blue}{CONTEXT}:
    \textit{maxlen}
    \newline
    \newline
    Impression summary: 
    \newline
     & You are an assistant designed to write impression summaries for the radiology  report. 
     \newline
     Users will send a context text and you will respond with an \newline impression summary using that context. 
    \newline
    \newline
    \textcolor{red}{Instructions:} 
    \begin{itemize}
        \item Impression should be based on the information that the user will send in the context.
        \item The impression should not mention anything about follow-up actions.
        \item Impression should not contain any mentions of prior or previous studies.
        \item Limit the generation to \textit{maxlen} words.
    \end{itemize}
    & \textcolor{blue}{CONTEXT}:
   \textit{context}
    \newline
    \newline
    Impression summary:  \\ 
    \bottomrule
  \end{tabular}
  \label{tab:Table 1} 
\end{table}

\subsection{Structured Report Output}

We also experiment with the ability to modulate the report generation output with specifications on the desired report output format in the prompts as few shot examples. It can be interesting for the radiologist and downstream applications to generate certain attributes of interest from the radiology report apart from generating the free text radiology impression. These attributes of interest could be extracting the pathologies, severity related to pathology, severity, size, or position of findings etc. We instruct LLM to generate the radiology report in a structured output format containing the impression summary and attributes of interest seen in the retrieved context. We provide specifications on the pathology we are interested in and other attributes of interest along with few shot prompts as shown in the prompt design in \autoref{tab:Table 2}

\begin{table}[]
  \centering 
  \caption{Prompts for the OpenAI LLMs for Structured Radiology Report Generation from the retrieved reports as context text in the Few-Shot setting. The text inside brackets in the prompts corresponds to the variables used while formatting the prompt.}
  \begin{tabular}{p{3in} p{3in}}
  \toprule
    \textbf{Prompt Design} & \textbf{Few Shots Example} \\
    \midrule
    Generate an impression summary for the radiology report using the context.
    \newline
    Pathology for impression should be from list of words as in: \{pathology\}
    \newline
    Positional words should be from list of words as in:  \{positional\_words\}
    \newline
    Severity should should be from list of words as in: \{severity\_words\}
    \newline
    Size should should be from list of words as in: \{size\_words\}
    \newline
    CONTEXT: \{example\_context\}  
    \newline
    IMPRESSION: \{example\_report\_json\}
    \newline
    CONTEXT: \{example\_context\}
    \newline
    IMPRESSION: \{example\_report\_json\}
    \newline
    CONTEXT: \{example\_context\}
    \newline
    IMPRESSION: \{example\_report\_json\}
    \newline
    CONTEXT: \{context\}
    \newline
    IMPRESSION: 
 & 
   CONTEXT: \newline
Right suprahilar opacities may relate to pulmonary vascular congestion although infectious process or aspiration not entirely excluded in the appropriate clinical setting. \newline

IMPRESSION:  \newline
\{  \newline
          "impression": "Mild bibasilar atelectasis is present. Right suprahilar opacities may be due to pulmonary vascular congestion. ",  \newline
          "attributes": [  
          \begin{quote}
                           \{  \newline
                          "pathology": "atelectasis",  \newline
                          "positional": "bibasilar"     \newline
                  \},   
           \end{quote}      
           \begin{quote}
                  \{     \newline
                          "pathology": "opacities",   \newline
                          "positional": "Right suprahilar”   \newline
                  \} ] 
            \end{quote} \}

\\ 
    \bottomrule
  \end{tabular}
  \label{tab:Table 2} 
\end{table}

\subsection{Experiments}

We conducted the retrieval augmented generation experiments using the OpenAI LLMs - \texttt{text-davinci-003}, \texttt{gpt-3.5-turbo} and \texttt{gpt-4} based on the retrieved records using the CXR-ReDonE embeddings. We consider both report-based corpus and sentence-based corpus in our experiments. The retrieved records from the corpus forms the context in the prompt based on which the LLM generates the free text radiology impression. We experimented with zero shot settings for free text impression generation and few shot settings for the structured report generation.

\section{Results on Real Data} 
\subsection{Evaluation Dataset}

We evaluate the performance on two golden benchmark datasets, MS-CXR [\cite{boeckingcxr}] and the test impressions from the  CXR-PRO dataset [\cite{rameshcxr}] . Both datasets are created with the help of board-certified radiologists. CXR-PRO consists of 2,188 radiology images and associated reports. CXR-PRO dataset was preprocessed to remove duplicate lines. MS-CXR dataset provides 1162 image–sentence pairs across eight different cardiopulmonary radiological findings. Both these datasets though from the same data distribution are different, MS-CXR is a phrase grounding dataset which also contains bounding boxes to ground the phrases on the radiology image, it contains very precise and concise phrases whereas CXR-PRO has longer text for the impressions. We evaluate the performance across these two setups as in some clinical settings it may be required to give precise and concise phrases about the radiology image. 

\subsection{Evaluation Approach}
We evaluate the free text radiology impressions generated using the LLMs from the retrieved records of the report level corpus and sentence level corpus. For sentence level corpus we evaluate the impressions generated from top K sentence retrievals with K= {1, 2, 3. Our baselines are the impressions retrieved from CXR-ReDonE. We evaluate on the two semantic metrics – BERTScore [\cite{zhang2019bertscore}] and $S_{emb}$ [\cite{endo2021retrieval}]  to see the similarity of the generation to the ground truth impression. We see this more meaningful as in the medical context phrases like lung collapse can represent atelectasis though the exact word may not be in the sentence. BERTScore computes a similarity score for each token in the predicted impressions with each token in the ground truth impressions. Token level similarity is computed using contextual embeddings instead of direct token matches. $S_{emb}$ uses CheXbert model [\cite{smit2020chexbert}] to calculate the cosine similarity between the embeddings from the final hidden state representations. To evaluate the overlap in clinical entities included in both the generated and ground truth reports we use RadGraph $F_{1}$, a metric proposed by \cite{yu2022evaluating} that makes use of a RadGraph model [\cite{jain2021radgraph}] to evaluate the overlap in clinical entities.

We also generate the radiology report impressions in a structured json format to evaluate if we can format the output the generated impression as per user requirements. At this point in time these are not quantitatively measured but we show qualitative outputs for structured generation.

\subsection{Measuring Hallucinations}
We also qualitatively and quantitatively evaluate if the report impression summary hallucinates from the top K sentences corpus given to the LLM as context. We use the $S_{emb}$ score between the LLM generations and the top-k context records to see if they semantically equivalent. 

\subsection{Results – CXR-PRO}
We evaluate the performance of radiology report impressions generated using a purely retrieval-based approach in CXR-ReDonE as our baseline. We find that RAG based generations improves the BERTScore metrics for both report and sentence corpus retrievals bringing in an absolute improvement of 5.06\% at k=3 for sentence-based retrieval. Similarly, it also improves $S_{emb}$ scores for both report and sentence corpus retrievals, bringing in an absolute improvement of 2.43\% at k=3 for sentence-based retrieval. RadGraph $F_{1}$ metric that measures the retrievals of clinical entities is on par with CXR-ReDonE at k=3 and slightly lower at lower k values. We should note that our approach generates the impressions on the augmented data so we cannot exceed CXR-ReDonE on RadGraph $F_{1}$ as otherwise we are hallucinating clinical entities not based on the context. The evaluation metrics are available in \autoref{tab:Table 3}.

\begin{table}[]
  \centering 
  \caption{Evaluation of CXR-RePaiR-Gen(RAG based approach) on CXR-PRO test impressions for report corpus retrieval and sentence corpus retrieval. Metrics evaluated are BERTScore, $S_{emb}$ score and RadGraph $F_{1}$. Our approach outperforms the baseline on both the clinical metrics BERTScore and $S_{emb}$ score for both the report and sentence corpus-based retrieval for all values of K and at par with CXR-ReDonE for RadGraph $F_{1}$ at k=3. Italics denote improvement over the baseline, bold denotes the highest value obtained by our approach at the highest value of RadGraph $F_{1}$ which is at k=3. We should note that CXR-RePaiR-Gen (based on RAG) cannot exceed CXR-ReDonE on the RadGraph $F_{1}$ score as otherwise it means we are hallucinating on the clinical entities which are not present in the context.}
  \begin{tabular}{lllll}
  \toprule
    \textbf{K} & \textbf{Method} & & \textbf{Evaluation Metrics} & \\
     & & \textbf{BERTScore} & \textbf{Semb} & \textbf{RadGraph F1} \\
    \midrule
    N/A & CXR-ReDonE & 0.2482 & 0.3647 & 0.0921 \\ 
    & CXR-RePaiR-Gen (text-davinci-003) & \textit{0.2600} & \textit{0.3741} & 0.0839 \\
    1 & CXR-ReDonE & 0.2455 & 0.4029 & 0.0861 \\
    & CXR-RePaiR-Gen (text-davinci-003) & \textit{0.2610} & \textit{0.4116} & 0.0774 \\
    2 & CXR-ReDonE & 0.2465 & 0.3892 & 0.1045 \\
    & CXR-RePaiR-Gen (text-davinci-003) & \textit{0.2753} & \textit{0.4036} & 0.0926 \\
    3 & CXR-ReDonE & 0.2276 & 0.3787 & 0.1104 \\
    & CXR-RePaiR-Gen(text-davinci-003) & \textit{0.2782} & \textit{0.4030} & 0.1018 \\
    & CXR-RePaiR-Gen(gpt-3.5-turbo) & \textit{0.2748} & \textit{0.3973} & 0.0991 \\
    & CXR-RePaiR-Gen(gpt-4) & \textbf{\textit{0.2865}} &  \textbf{\textit{0.4026}} &  \textbf{0.1061} \\
    \bottomrule
  \end{tabular}
  \label{tab:Table 3} 
\end{table}

\subsection{Results – MS-CXR}

Our approach improves the BERTScore on MS-CXR phrase grounding text by an absolute value of 8.67 and $S_{emb}$ by an absolute value of 3.86. The evaluation metrics are available in \autoref{tab:Table 4}.

\begin{table}[]
  \centering 
  \caption{Evaluation of CXR-RePaiR-Gen on MS-CXR phrases using text-davinci-003 for sentence corpus-based retrieval at k = 3. Metrics evaluated are BERTScore, $S_{emb}$ and RadGraph $F_{1}$. Our approach outperforms the baseline on both the clinical metrics BERTScore and $S_{emb}$ and at par for RadGraph $F_{1}$. Italics denotes improvement over the baseline, while bold denotes the highest value obtained by our approach. }
  \begin{tabular}{llll}
  \toprule
    \textbf{Method} & & \textbf{Evaluation Metrics} & \\
     & \textbf{BERTScore} & \textbf{Semb} & \textbf{RadGraph F1} \\
    \midrule
    CXR-ReDonE & 0.1102 & 0.3494 & 0.0626 \\ 
    CXR-RePaiR-Gen \newline (text-davinci-003) & \textbf{\textit{0.1970}} & \textbf{\textit{0.3880}} & \textbf{0.0617} \\ 
 
    \bottomrule
  \end{tabular}
  \label{tab:Table 4} 
\end{table}

\subsection{Qualitative Analysis}

We find the retrieval augmented generations-based impressions are very concise and less noisy when compared to the outputs from a pure retrieval-based strategy but still retaining all the relevant clinical entities. Refer \autoref{tab:Table 5} and \autoref{tab:Table 6} to see the concise impression summary created by the RAG based approach for examples from CXR-PRO and MS-CXR respectively. RAG also tries to avoid insignificant details in the retrieved corpus. In the last sample of \autoref{tab:Table 5}, RAG based generations automatically exclude details like follow-up recommendations and gives a cleaner impression.

% \begin{table}[]
%   \centering 
%   \caption{Prompts for the Open AI LLMs for Radiology Report Impression Generation from the retrieved reports as context text in the zero-shot setting. The text in italics in the prompts corresponds to the variables used in formatting the prompt.}
  \begin{longtable}{p{1.5in} p{1.5in} p{1.5in} p{1.5in}}
  \caption{Examples from CXR-PRO dataset: Generated reports compared to the reference report and the baseline method CXR-RedonE. RAG based generations from CXR-RepaiR-Gen provides very concise impression summaries with ability to retain the important clinical entities. Outputs were generated at top K retrievals set to 3.}\\  
  \toprule
    \textbf{Ground Truth} & \textbf{CXR-ReDonE (K=3)} & \textbf{CXR-RepaiR-Gen (K=3) \newline (text-davinci-003)} & \textbf{CXR-RepaiR-Gen (K=3) \newline (gpt-4)} \\
    \midrule
    \textcolor{red}{Right hilar mass} is associated with atelectasis of right middle lobe and there is a significant adjacent lymphangitic carcinomatosis and \textcolor{blue}{consolidation}. Right hilar mass is associated with \textcolor{orange}{atelectasis} of \textcolor{blue}{right middle lobe} and there is a significant adjacent \textcolor{green}{lymphangitic carcinomatosis} and consolidation. & 
    Although this may reflect dense consolidative pneumonia, findings at least raise concern for the possibility of an underlying \textcolor{red}{mass} with associated pneumonitis as a possible alternative diagnosis. Regions of confluent \textcolor{blue}{consolidation in the right middle lobe} and right suprahilar region, which could represent superimposed infection or \textcolor{orange}{atelectasis}. Right middle lobe consolidation in part due to known underlying \textcolor{green}{metastatic} disease with possible superimposed postobstructive infection or atelectasis. &
    Large \textcolor{red}{right perihilar mass} with associated \textcolor{orange}{extensive atelectasis} and regions of confluent \textcolor{blue}{consolidation in the right middle lobe} and right suprahilar region, raising concern for underlying \textcolor{green}{malignancy} and possible alternative diagnosis of pneumonitis. &
    Apparent \textcolor{red}{right hilar mass}, concerning for obstructive adenopathy. Confluent \textcolor{blue}{consolidation in right middle lobe} and suprahilar region, possibly due to infection or atelectasis. Findings worrisome for large right perihilar mass with \textcolor{orange}{extensive atelectasis}, suggesting potential \textcolor{green}{malignancy}. \\
    PA and lateral chest: Significantly low lung volume exaggerates mild \textcolor{red}{cardiomegaly}, but pulmonary vascular engorgement accompanied by mild \textcolor{orange}{interstitial edema} in the lower lungs has an indication of \textcolor{green}{cardiac decompensation}. &
    Clearing failure.Moderate \textcolor{orange}{pulmonary edema} from the right upper lung , and residual edema considered as the likely explanation for residual opacification at the right lung base , unless it proves subsequently unresponsive to diuresis.Mild \textcolor{orange}{pulmonary edema} with moderate \textcolor{red}{cardiomegaly} , small right pleural effusion , and cephalization of vasculature. &
    Mild pulmonary edema with moderate \textcolor{red}{cardiomegaly}, small \textcolor{red}{bilateral pleural effusions}, and cephalization of vasculature, reflecting \textcolor{green}{cardiac decompensation}. &
    Mild \textcolor{orange}{pulmonary edema}, moderate \textcolor{red}{cardiomegaly}, small right pleural effusion, cephalization of vasculature, and possible infectious process in right perihilar opacities. \\
    AP chest: Lung volumes are lower, exaggerating severe \textcolor{blue}{enlargement of the cardiac silhouette} and volume of moderate-sized \textcolor{red}{bilateral pleural effusions}.There is enlargement of the cardiac silhouette with \textcolor{orange}{pulmonary vascular congestion} and bilateral opacifications consistent with significant layering effusions and \textcolor{green}{compressive basilar atelectasis}. &
    Hazy opacification of the hemithoraces is with \textcolor{red}{bilateral pleural effusions} and compressive atelectasis , along with \textcolor{blue}{enlargement of the cardiac silhouette} and \textcolor{orange}{pulmonary vascular congestion}. Hazy opacification of the lower lungs suggests bilateral layering pleural effusions with some atelectatic changes at the bases.Hazy opacifications bilaterally are consistent with pleural effusions and \textcolor{orange}{compressive atelectasis at the bases}. &
    \textcolor{red}{Bilateral pleural effusions}, \textcolor{green}{compressive atelectasis}, \textcolor{blue}{enlargement of cardiac silhouette}, and \textcolor{orange}{pulmonary vascular congestion} are seen. Mild pulmonary edema and interstitial edema cannot be excluded. &
    \textcolor{red}{Bilateral hazy opacifications, layering pleural effusion}, \textcolor{green}{compressive atelectasis}, \textcolor{blue}{enlarged cardiac silhouette}, \textcolor{orange}{pulmonary vascular congestion}, bibasilar opacities, and possible mild pulmonary edema. \\   
    \bottomrule
    \label{tab:Table 5} 
  \end{longtable}

\begin{table}[]
  \centering 
  \caption{Examples from MS-CXR dataset: Generated reports compared to the reference report and the baseline method CXR-RedonE. RAG based generations from CXR-RepaiR-Gen avoids noisy and duplicate details in its generation and still retains the important clinical entities. Outputs were generated at top K retrievals set to 3.}
  \begin{tabular}{p{2in} p{2in} p{2in}}
  \toprule
    \textbf{Ground Truth} & \textbf{CXR-ReDonE (K=3)} & \textbf{CXR-RepaiR-Gen (K=3)\newline(text-davinci-003)} \\
    \midrule
    Subtle opacity in the left perihilar region & 
    \textbf{Vague asymmetric opacity in the left lower lobe which may represent an area of early infection. Vague left infrahilar density , } possibly associated with lower airway inflammation or infection , or potentially early or mild bronchopneumonia , but not of definite significance.\textbf{Indeterminate 3 cm ovoid density} in the left lung base with possible retrocardiac correlate on lateral view. & Vague asymmetric opacity in the left lower lobe is suggestive of early infection or developing pneumonia, depending on the clinical scenario. \\ \\
    Left lower lobe consolidation due to pneumonia & More confluent basilar opacities may reflect dependent edema , but aspiration and infectious pneumonia should also be considered in the appropriate clinical setting.Retrocardiac opacity likely reflects basilar atelectasis , however infection or aspiration may produce a similar appearance , \textbf{and short term followup radiographs may be helpful in this regard if warranted clinically.Right internal jugular line tip is in the right atrium and is needs to be pulled back to be indicate labral junction or above , 3. 5 cm have to be pulled back.} & Mild pulmonary edema with patchy opacities and bilateral effusions, predominantly on the left. Compressive atelectasis and pneumonia are possible. \\
    \bottomrule
  \end{tabular}
  \label{tab:Table 6} 
\end{table}

\subsection{Instruction Driven Output}

We also evaluate the instructions following capabilities of GPT-4 by passing specific prompts to exclude any mentions of prior report reference and follow-up actions in the generated impression. \newline

\textbf{Observations from CXR-PRO evaluation:}

\begin{itemize}
    \item All the 40 occurrences related to follow-up recommendations in the retrieved context documents were not included in the final impression. 
    \item Out of 87 references to previous/prior reports, 80 references were successfully excluded in the impression.
\end{itemize}

Some sample cases where the prior report references and follow-up actions were removed based on the instructions in the prompt are shown in \autoref{tab:Table 7}

\begin{table}[]
  \centering 
  \caption{Random examples of cases from CXR-PRO dataset where the RAG based generations from GPT-4 eliminated the references to prior reports and follow-up recommendations based on instruction-based prompts.}
  \begin{tabular}{p{3in} p{3in}}
  \toprule
    \textbf{Context Records} & \textbf{RAG Impression}\\
    \midrule
    Differential appearance include interstitial lung disease with or without superimposed developing pneumonia versusasis associated with elevation of the right hemidiaphragm.Patchy right basilar opacity may reflect compressive atelectasis however infection cannot be excluded and \textbf{followup radiographs after diuresis are recommended.} Bibasilar hazy opacities , right greater than left , possibly due to atelectasis given low lung volumes , however underlying infection cannot be excluded in the appropriate clinical setting. &
    Interstitial lung disease with possible developing pneumonia or atelectasis, patchy right basilar opacity, and bibasilar hazy opacities, right greater than left. Infection cannot be excluded. \\ \\
    It is unusual to have a unilateral left effusion \textbf{when the patient has previously had right sided effusions and causes such as pericarditis, pancreatitis or a left upper quandrant process should be considered.} Bilateral opacifications with silhouetting hemidiaphragms is consistent with layering pleural effusions and underlying compressive basilar atelectasis. Bilateral layering effusions are present with patchy airspace opacity predominantly at the bases, left greater than right, which may represent patchy atelectasis. &
    Unilateral left effusion, bilateral opacifications with silhouetting hemidiaphragms, compressive basilar atelectasis, and patchy airspace opacity predominantly at the bases, left greater than right.\\
    \bottomrule
  \end{tabular}
  \label{tab:Table 7} 
\end{table}

\subsection{Modulate Report Generation Format}

One of the key advantages of RAG is the ability to modulate output generation format via prompt engineering. \autoref{tab:Table 8} shows the report generation outputs in a structured json format extracting the attributes of interest using the few shot prompts in \autoref{tab:Table 2}.

% TODO TABLE 8

\begin{table}[]
  \centering 
  \caption{Structured Report Generation outputs from CXR-RepaiR-Gen using the few shot prompts in \autoref{tab:Table 2}. We see that we can generate the impression summary and extract pathology, positional information, severity, size from the retrieved context in a structured format. }
  \begin{tabular}{p{6in}}
  \toprule
    \textbf{Structured Report Output (JSON)} \\

    \midrule
    \{ \newline
   "impression":"The Swan-Ganz catheter tip is seen in the proximal right pulmonary artery. Appropriate position of Dobbhoff line reaching stomach. Combination of severe bilateral lower lobe atelectasis and small to moderate pleural effusions” \newline

   "findings":[  
     \begin{quote}
     \{

         "pathology":"atelectasis", \newline
         "positional":"bilateral, base", \newline
         "severity":"severe", \newline
         "size":"" \newline
      \}, 
      \end{quote}
     \begin{quote}
      \{ \newline
         "pathology":"pleural effusions", \newline
         "positional":"bilateral", \newline
         "severity":"", \newline
         "size":" small to moderate" \newline
      \}
       \end{quote}
       ] 
   \}
\\

    \bottomrule
  \end{tabular}
  \label{tab:Table 8} 
\end{table}

\subsection{Hallucinations in Retrieval Augmented Generation}

We calculate the $S_{emb}$ scores between the generated report impression and the retrieved context from the sentence corpus with K=3 top K retrievals. We find that the average similarity score is 0.8466 and around 87\% of the test impressions have a $S_{emb} > 0.70$ which is a good indication that the generations did not hallucinate.  \autoref{tab:Table 9} presents a couple of records which had the lowest $S_{emb}$ scores in the test set. We find that even such records did not hallucinate, and the lowest scores may be attributed to the concise impression summary generated.

\begin{table}[]
  \centering 
  \caption{Examples of cases where the Semb scores were less between the retrieved context sentence corpus and the generations from the LLM model. Semb uses CheXbert model to calculate the cosine similarity between the embeddings from the final hidden state representations and can give an indication of hallucination. We see there are no hallucinations.}
  \begin{tabular}{p{2in} p{2in} p{2in}}
  \toprule
    \textbf{$S_{emb}$} & \textbf{Context Records (K=3)} & \textbf{RAG Impression} \\
    \midrule
    0.1508 & 
    Lung volumes with right basilar atelectasis versus consolidation which could represent pneumonia in the right clinical context. Low lung volumes with bibasilar opacities which may represent atelectasis versus aspiration. Low lung volumes with bibasilar opacities which could potentially be due to atelectasis. &
    Right basilar atelectasis and consolidation likely due to pneumonia. Bibasilar opacities may be due to atelectasis or aspiration. Low lung volumes present. \\
    0.2116 &
    The left lung remains grossly clear, although there is minimal linear opacity at the left costophrenic angle, which may represent post - inflammatory scarring or subsegmental atelectasis. There is a suggestion of a tiny left apical pneumothorax which me if there is attempted line placement from the left side prior to the right. There is enlargement of the cardiac silhouette there is suggestion of some mild engorgement of pulmonary vessels on the left, raising the possibility of asymmetric elevation of pulmonary venous pressure. &
    The left lung is clear with minimal linear opacity at the left costophrenic angle. Tiny left apical pneumothorax and mild engorgement of pulmonary vessels on the left suggest asymmetric elevation of pulmonary venous pressure. Enlargement of cardiac silhouette noted. \\
    \bottomrule
  \end{tabular}
  \label{tab:Table 9} 
\end{table}

\section{Discussion} 
We find that CXR-RepaiR-Gen - Retrieval Augmented Generation based approach for radiology report impression generation using embeddings from contrastively pretrained model CXR-ReDonE for retrieval of relevant radiology text and OpenAI LLM models for generation can help generate very concise and precise radiology report impressions and improve the clinical efficacy metrics specifically the BERTScore and $S_{emb}$ scores without compromising on retaining the relevant clinical entities. We also find that the LLM based report generations via few shot prompts and instructions allow for modulation of report outputs for desired content and format and helps reduce noisier text (like recommendations for further evaluation, prior report mentions, specific patient details etc) and avoids repeating content in the impressions which are the typical issues with a pure retrieval-based setup.  We note that these limitations were mentioned in the CXR-RePaiR paper which warranted the requirement to have a template database that removed these noisy details or duplicates and use the template database for performing the retrievals for 
better quality report generation. We see that RAG based generation can address this limitation via effective prompt engineering and also using approaches like structured content extraction with few shot prompts to extract only the attributes of interest and produce a more concise, precise, and complete impression summary from the retrieved records.

\subsection{Limitations}
We note that as RAG impression generation efficacy is sensitive to the retrieved sentences from the corpus which is based on the embeddings from the constrastively pretrained model meaning that the clinical entities generated by RAG is limited by the clinical entities from the retrieval unless we allow it to use the general knowledge it has. In our experiments we restricted the model to use the context while making the generations via prompt instructions to alleviate hallucinations. So it is imperative that the retrieval model is able to bring in the all the relevant clinical entities for the generation.

\subsection{Future Work}
We see that RAG based report generations can benefit from more advanced contrastive models that are more sensitive to fine details of the radiology image such as severity, size, position, pathology, and other attributes of interest from the radiology image. Advances in prompt engineering for medical text is another area to explore so that we can elicit the LLM more efficiently for a specific downstream task.

% ACKNOWLEDGEMENTS ONLY GO IN THE CAMERA-READY, NOT THE SUBMISSION
% \acks{Many thanks to all collaborators and funders!}

\section{Appendix}
\subsection{GPT-3 Hyperparameters}
We tried with temperature values of {0, 0.5 and 1} for the zero shot impression generations from the text-davinci-003 model using the top K retrievals(K=3) from the sentence level corpus as the context. We found that it did not influence the output scores much as seen from \autoref{tab:Table 10}. The instruction to use retrieved sentence corpus as context for generations may be responsible for this behavior. 

\begin{table}[]
  \centering 
  \caption{RAG based report generation metrics at different temperature settings for the text-davinci-003 model using the sentence corpus and top K retrievals set to K=3.}
  \begin{tabular}{llll}
  \toprule
    \textbf{Temperature} & & \textbf{Evaluation Metrics} & \\
     & \textbf{BERTScore} & \textbf{Semb} & \textbf{RadGraph F1} \\
    \midrule
    T=0 & 0.2782 & 0.4030 & 0.1018 \\ 
    T=1 & 0.2783 & 0.3996 & 0.1013 \\ 
    T=2 & 0.2787 & 0.3998 & 0.1011 \\ 
    \bottomrule
  \end{tabular}
  \label{tab:Table 10} 
\end{table}

\bibliography{ms}

% \section{Appendix}
% \input{content/appendix}
\end{document}